# Top k Memory Candidates in Memory Networks for Common Sense Reasoning

Vatsal Mahajan

vmahaja1@asu.edu

**Abstract**—Successful completion of reasoning task requires the agent to have relevant prior knowledge or some given context of the world dynamics. Usually, the information provided to the system for a reasoning task is just the query or some supporting story, which is often not enough for common reasoning tasks. The goal here is that, if the information provided along the question is not sufficient to correctly answer the question, the model should choose k most relevant documents that can aid its inference process. In this work, the model dynamically selects top k most relevant memory candidates that can be used to successfully solve reasoning tasks. Experiments were conducted on a subset of Winograd Schema Challenge (WSC) problems to show that the proposed model has the potential for commonsense reasoning. The WSC is a test of machine intelligence, designed to be an improvement on the Turing test.

## 1 INTRODUCTION

THERE are lots of AI tasks that require (dynamic) knowledge from external sources to accomplish a task. Such as in case of NLU, deep QA, and hard co-reference resolution. Memory Networks help tackle this problem. Memory Networks combine the successful learning strategies developed in the machine learning literature for inference with a memory component that can be read and written to. The model is then trained to learn how to operate effectively with the memory component [Weston *et al.*, 2015]. Memory Networks use memory vectors to provide relevant information along with the input query for the learning and inference process. These memory slots are defined by the data provided along with input as in case of Facebook bAbl tasks - a question and a story along with it.

In this work, I use a simple technique to dynamically select top k candidates for memory slots, for any given task. This is useful in the case where the input query does not have any story or additional supporting data along with it. I hypothesize that this additional information would aid a model's inference process in such uses cases.

## 2 RELATED WORK

[Miller *et al.*, 2016] use a similar technique to create Key-Value Memory Networks for directly reading form documents. They use the question to preselect a small subset of the possibly large array of relevant info. But, he tasks designed here have both the questions and the supporting document. The tasks here are not designed to test the system's commonsense reasoning ability. Whereas, this work aims at inferring the cause-effect relationship between pairs of events by dynamically selecting memory candidates to support the inference. [Liu et al., 2016] use Neural Association models to learn cause-effect pairs. They base their work on collecting the cause-effect relationships between a set of common words and phrases. They also hypothesize that this type of knowledge would be a key component for modeling the association relationships between discrete events.

## 3 WINOGRAD SCHEMA

The Winograd Schema (WS) evaluates a system's commonsense reasoning ability based on a traditional, very difficult natural language processing task: co-reference resolution [Levesque et al., 2011]. It is carefully designed such that a task cannot be easily solved without commonsense knowledge [Liu *et al.*, 2016]. Winograd Schema Challenge (WSC) poses a set of multiple-choice questions that have a particular form for example:

*Sentence: The trophy would not fit in the brown suitcase because it was too big (small).*
*Question: What was too big (small)?*
*Answer0: the trophy Answer1: the suitcase*

To answer the above question, one requires having the knowledge that an object being big would have a higher chance of not fitting in a suitcase, as compared to a small object [Mahajan, 2018]. Some external knowledge is required to help with this spatial reasoning.

In the first experiment, I am working towards the Winograd Schema Challenge. The goal here is to evaluate the proposed model for various reasoning tasks. Here the model is trained to infer the cause-effect relationship between discrete events.

## 4 APPROACH

### 4.1 Memory Nets

I have used the End-to-End Memory Net described by [Weston *et al.*, 2015]. The model is defined by a discrete set of inputs $x_1, x_2, ..., x_n$ that are to be stored in the memory, a query $q$, and outputs an answer $a$. All the symbols in each of the $x_i, q$ and $a$ come from a vocabulary $V$ of words. The End-to-End Memory Network model writes all the inputs $x$ to the memory up to a fixed buffer size, and then finds a continuous representation for the $x$ and $q$.

Using an embedding matrix $A$ of size $d \times V$, the set of $\{x_i\}$ is converted into memory vectors $\{m_i\}$ of dimension $d$ computed by embedding each $x_i$ in a continuous space. The query $q$ is also embedded using another embedding matrix $B$ of size $d \times V$, to obtain an internal state $u$. In the embedding space, the model computes the match between $u$ and each memory $x_i$ by taking the inner product followed by a softmax: $p_i = Softmax(u^T.m_i)$. Once the model is trained $p$ gives the probability vector over the memory slots. Also, each $x_i$ has a corresponding output vector $c_i$ generated by another embedding matrix $C$ of size $d \times V$. Then the response vector $o$ is computed from the memory which is given by the sum over the transformed inputs $c_i$, weighted by the probability vector from the input: $o = \sum_i p_i . c_i$ . To get the final prediction the sum of the output vector $o$ and the input embedding u is then passed through a final weight matrix $W$ (of size $V \times d$) and a softmax to produce the predicted label: $\hat{a} = Softmax(W(o + u))$. During training, all three embedding matrices $A$, $B$, and $W$, as well as $W$ are jointly learned by minimizing a standard cross-entropy loss between $\hat{a}$ and the true label $a$. Here, I have discussed a single layer model, but this can also be stacked to get multiple hops in memory.

### 4.2 Indexing Corpora

The goal here is to index a corpus so that when given a query/question the system can pick top $k$ memory candidate. The objective is to get the set of the $k$ most relevant memory candidates given a particular query string $q$. More formally, let $Q \subseteq \mathbb{R}^n$ be an input space, $W \subseteq \mathbb{R}^m$ be a finite output space of size N, and $f : Q \times W \rightarrow \mathbb{R}$ a known scoring function [Goel $et\ al.$, 2008]. Given an input (search query) $q \in Q$, the goal is to find, or closely approximate, the top-k output objects (memory candidates) $z_1, z_2, \dots, z_k$ in $W$ (i.e., the top $k$ objects as ranked by $f(q, \cdot)$. The most commonly used indexing function is the inverted index. [Zobel $et\ al.$ 2006] give a detailed description of inverted indexing.

### 4.3 Dynamically Selecting Memory Candidates

For reasoning tasks where, one has the input as just a query/question $q$ and no supporting story/data along with it, the indexed corpora can be used for generating top $k$ candidates for memory slots. The candidates for the memory slots will be generated dynamically. The quality of the selected candidates will only depend on the type of scoring function $f$ chosen.

Now the discrete set of inputs $x_1, x_2, \dots, x_n$ that are to be stored in the memory are chosen using the function $f(q, \cdot)$ from the corpora $W$. So now $\{x_i\} = \{z_i\}$; where $z_i$ is one of the k candidates given by the function $f$. Further, I use the same embedding process to convert the set $\{z_i\}$ into memory vectors $\{m_i\}$.

### 4.3 Creating Cause-Effect Pairs

For the experiment, I have used the wiki corpus for selecting memory candidates. And Causal Time bank to get cause-effect pairs for training. The Causal Time bank [Mirza et al., 2014] has 318 causal links (event pairs). So, I created a data set of cause-effect pairs with 318 positive examples and 954 negative examples using the causal time bank. For testing, I labeled 62 cause-effect problems from a total of 278 available WS questions for this experiment.

The WSC questions were mapped to a cause effect problem. Each input question was mapped to case-effect query as "A <CAUSES> B", where A and B are actions/verbs and <CAUSES> is special token denoting a causal event/verb.

### 4.3 Model

The memory network is trained to output the Probability (A causes B). For embedding the input query $q$, I have used position encoding (PE), meaning that the order of the words now affects $m_i$ [Sukhbaata et al., 2015]. This representation that encodes the position of words within the sentence takes the form: $m_i = \sum_j l_j . Ax_{ij}$ , where · is an element-wise multiplication. $l_j$ is a 4-column vector with the structure $l_{kj} = (1 - j/l) - (k/d)(1 - 2j/J)$ (assuming 1-based indexing), with $J$ being the number of words in the sentence, and $d$ is the dimension of the embedding. The same representation is used for memory inputs and memory outputs. The network was trained with 2 memory hops. The model was trained using Adam optimizer.

## 5 RESULTS

Out of the 62 questions, the trained model was able to correctly answer 25 questions. The network currently overfits to the training data due to the small size to the training data.

**Future Work**: [Liu et al., 2016] describe an approach to extract cause-effect pairs, estimated to be around 500,000 pairs. The data set is called CauseCom, but it is not available publically. Currently, I am working towards recreating the data-set as described in their paper. With more data to train the network, the results should show an improvment over the current accuracy.